\documentclass{ecai}
\usepackage{times}
\usepackage{graphicx}
\usepackage{latexsym}

\usepackage{url}            
\usepackage{booktabs}       
\usepackage{amsfonts}       
\usepackage{nicefrac}       
\usepackage{microtype}      
\usepackage{amsmath}
\usepackage{subcaption}
\usepackage{makecell}

\ecaisubmission   

\begin{document}

\title{Unsupervised Learning of Anomaly Detection from Contaminated Image Data using Simultaneous Encoder Training}

\author{Amanda Berg$^{1,2}$, J\"orgen Ahlberg$^{1,2}$, Michael Felsberg$^{2}$\\
$^1$Termisk Systemteknik AB, Diskettgatan 11 B, 583 35 Link\"oping, Sweden\\
$^2$Computer Vision Laboratory, Dept. EE, Link\"oping University, 581 83 Link\"oping, Sweden\\
{\tt\small \{amanda.,jorgen.ahl\}berg$@$termisk.se, \{amanda.,jorgen.ahl,michael.fels\}berg$@$liu.se}}

\maketitle
\bibliographystyle{ecai}

\begin{abstract}
Unsupervised learning of anomaly detection in high-dimensional data, such as images, is a challenging problem recently subject to intense research. Through careful modelling of the data distribution of normal samples, it is possible to detect deviant samples, so called anomalies.  Generative Adversarial Networks (GANs) can model the highly complex, high-dimensional data distribution of normal image samples, and have shown to be a suitable approach to the problem. Previously published GAN-based anomaly detection methods often assume that anomaly-free data is available for training. However, this assumption is not valid in most real-life scenarios, a.k.a. in the wild. In this work, we evaluate the effects of anomaly contaminations in the training data on state-of-the-art GAN-based anomaly detection methods. As expected, detection performance deteriorates. To address this performance drop, we propose to add an additional encoder network already at training time and show that joint generator-encoder training stratifies the latent space, mitigating the problem with contaminated data. We show experimentally that the norm of a query image in this stratified latent space becomes a highly significant cue to discriminate anomalies from normal data. The proposed method achieves state-of-the-art performance on CIFAR-10 as well as on a large, previously untested dataset with cell images.
\end{abstract}

\section{Introduction}
Anomaly detection is the identification of \textit{rare} samples, objects, or events that are regarded as anomalous compared to what is considered to be normal. Anomalies are sometimes also referred to as outliers \cite{Hodge2004}. Due to the quite general problem formulation, anomaly detection is applicable to a wide range of different fields, such as e.g. agriculture \cite{Christiansen2016}, medicine \cite{Schlegl2017,Schlegl2019}, and finance \cite{Abdallah2016,Ahmed2016}. In the context of machine learning, anomaly detection can be \textit{supervised}, \textit{semi-supervised}, or \textit{unsupervised}. This paper addresses \textit{unsupervised} anomaly detection.

The objective of unsupervised anomaly detection is to detect previously unseen rare objects or events without any prior knowledge about these. The only information available is that the percentage of anomalies in the dataset is small, usually less than 1\%. Since anomalies are rare and unknown to the user at training time, anomaly detection in most cases boils down to the problem of modelling the normal data distribution and defining a measurement in this space in order to classify samples as anomalous or normal. In high-dimensional data such as images, distances in the original space quickly lose descriptive power (curse of dimensionality) and a mapping to some more suitable space is required. Due to their latent space, Generative Adversarial Networks (GANs) \cite{Goodfellow2014} can model complex, high-dimensional data distributions \cite{Creswell2017} and are, therefore, suitable for anomaly detection in images. GAN-based methods also provide the ability to localize anomalies within images in contrast to many classical anomaly detection methods \cite{Schlegl2017,Schlegl2019}. Although partly addressed in recent works  \cite{Akcay2019a,Akcay2019,Deecke2018,Ngo2019,Schlegl2019,Schlegl2017,Zenati2018a,Zenati2018}, unsupervised anomaly detection still remains a challenging problem. 

The main limitation of these previously published unsupervised GAN-based methods is their assumption that anomaly-free data is available for training. For this reason, we argue that they are not \textit{truly} unsupervised, since completely anomaly-free data requires weak labelling. Anomaly contamination of GAN training data is expected to reduce detection performance \cite{Beggel2019}. In this work, we show that this is indeed the case for a recent, state-of-the-art GAN based anomaly detection method f-AnoGAN \cite{Schlegl2019} and its variations. 

Further, we show using t-SNE visualization \cite{Maaten2008} that anomalous and normal validation samples are scattered in latent space such that the GANs expressiveness with respect to classification is limited. To mitigate this problem, an image-to-latent-space encoder trained \textit{jointly} with the generator is proposed. The joint training coupled with an image distance encoder loss enforces similar images to lie close to each other also in latent space. In this stratified latent space, latent vectors of anomalous samples prove to have shorter norms than those of normal samples. We show this empirically in a number of experiments on two datasets, based on CIFAR-10 and on a large cell-image dataset. Our approach achieves state-of-the-art performance in both cases.\\

\textbf{Contributions}
\begin{itemize}
\item We conduct an empirical study varying the amount of anomalies in the training data and measure the degradation of the anomaly detection in existing methods.
\item We propose an approach to truly unsupervised anomaly detection based on simultaneous encoder training that improves results even when the training data is contaminated with anomalies.
\end{itemize}

\section{Related work}
Anomaly detection is an important problem relevant to a vast number of fields, e.g. malware intrusion detection \cite{Kwon2017}, retinal damage detection \cite{Schlegl2019,Schlegl2017}, and detection of anomalous events in surveillance videos \cite{Sultani2018}. A complete review of anomaly detection methods is beyond the scope of this paper, the interested reader is referred to \cite{Chalapathy2019,Chandola2009}. In the particular case of \textit{unsupervised} anomaly detection, labels are unknown at training time. This paper is focused on unsupervised deep learning based anomaly detection of/in high-dimensional, non-sequential data with spatial coherence, i.e., images.

Classical methods for unsupervised anomaly detection include probabilistic methods that model the data distribution, e.g., by using a non-parametric Kernel Density Estimator (KDE) \cite{Parzen1962} as in \cite{Yeung2002} where it is applied to intrusion detection. Samples in low density areas are treated as anomalies. Another example of a probabilistic, parametric method is the RX anomaly detector \cite{Reed1990}. Due to the \textit{curse of dimensionality}, probabilistic methods are, however, not suitable for high-dimensional data such as images. Also, they typically do not provide the ability to \textit{localize} anomalies in images.

In contrast, reconstruction-based methods provide the possibility to localize anomalies within images. The aim of these methods is to find a lower-dimensional latent space from which normal samples can be reconstructed. A query image is then projected onto this latent space and the reconstructed image is compared to the query image by some image distance measurement in order to discriminate anomalous cases. The latent space can be modelled using, e.g., Auto Encoders \cite{Xia2015}, Variational Auto Encoders \cite{An2015}, or Generative Adversarial Networks (GANs) \cite{Akcay2019,Deecke2018,Schlegl2019,Schlegl2017,Zenati2018a,Zenati2018}. In the context of unsupervised anomaly detection, GANs were first introduced by Schlegl et. al. \cite{Schlegl2017} (AnoGAN). They proposed to use a combination of the $l_2$-norm and a discrimination loss between a query image and its closest reconstruction match as an anomaly score. Based on this approach, Deecke et. al. \cite{Deecke2018}  proposed a similar method (ADGAN) that improved the results slightly. In contrast to AnoGAN, ADGAN initialized the search in latent space for the closest match at multiple locations. Recently, and concurrent to this work, Schlegl et. al. \cite{Schlegl2019} proposed f-AnoGAN, improving their method (AnoGAN) by replacing the Deep Convolutional GAN (DCGAN) \cite{Radford2015} with a Wasserstein GAN (WGAN-GP) \cite{Gulrajani2017} and they also introduced an encoder that was trained separately for image to latent space mapping. The usage of an encoder instead of an iterative optimization procedure in order to speed up image to latent space mapping has also been explored by Zenati et. al. \cite{Zenati2018a,Zenati2018} who employed an architecture similar to a Bidirectional GAN (BiGAN) \cite{Donahue2016} with pairs of $(X,z)$ as input to the discriminator. We argue that the novelty of the proposed method compared to \cite{Zenati2018a,Zenati2018} is the discussion
of the impact of such an encoder on the structure of the latent space, and also the problem of training data contamination.

Ngo et al. \cite{Ngo2019} make the observation that the usual GAN objective encourages the distribution of generated samples to overlap with the real data, which may not be optimal in the case of anomaly detection. They further propose an \textit{encirclement} loss that places generated samples at the boundary of the distribution and can then use the discriminator directly to discriminate anomalous samples.

Golan and El-Yaniv \cite{Golan2018} proposed another type of method trained to map input images to a set of geometric transformations. In contrast to the reconstruction-based methods, it can not provide anomaly localization in images.

Some of the methods mentioned above \cite{Akcay2019,Deecke2018,Schlegl2019,Schlegl2017} claim to be unsupervised while at the same time assuming anomaly-free data for training. The acquisition of anomaly-free data requires labelling of data as normal. However, anomalous objects and/or events are rare and difficult to label in most real-world scenarios.

Beggel et. al. \cite{Beggel2019} conclude that the anomaly detection performance is reduced when the training set is contaminated with anomalies. They use an Adversarial Auto Encoder \cite{Makhzani2016} to mitigate the problem by rejecting potential anomalies already during training. The proposed method improves detection results in the case of anomalies present in the training data in a different way. Instead of rejecting, we propose to use an encoder trained jointly with the GAN. As we show in our experiments, the anomalies need not to be rejected at training time, but mapped closer to the origin.

\section{Method}
\begin{figure*}[t]
  \centering
  \begin{minipage}[c]{1\columnwidth}
    \centering 
    \includegraphics[width=1\columnwidth,trim={0 10.0cm 10cm 0},clip]{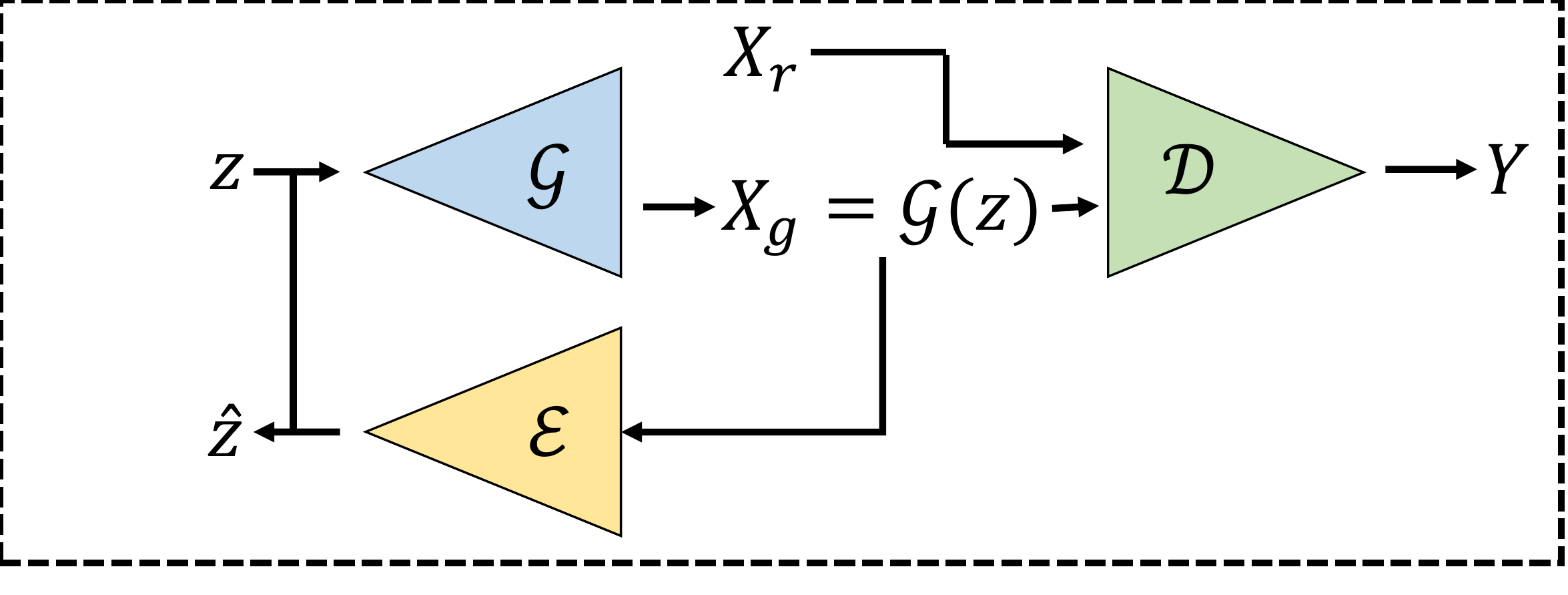}
    \subcaption{Training}\label{fig:overview_train}
  \end{minipage}
  \begin{minipage}[c]{1\columnwidth}
    \centering 
    \includegraphics[width=1\columnwidth,trim={0 10.0cm 10.0cm 0},clip]{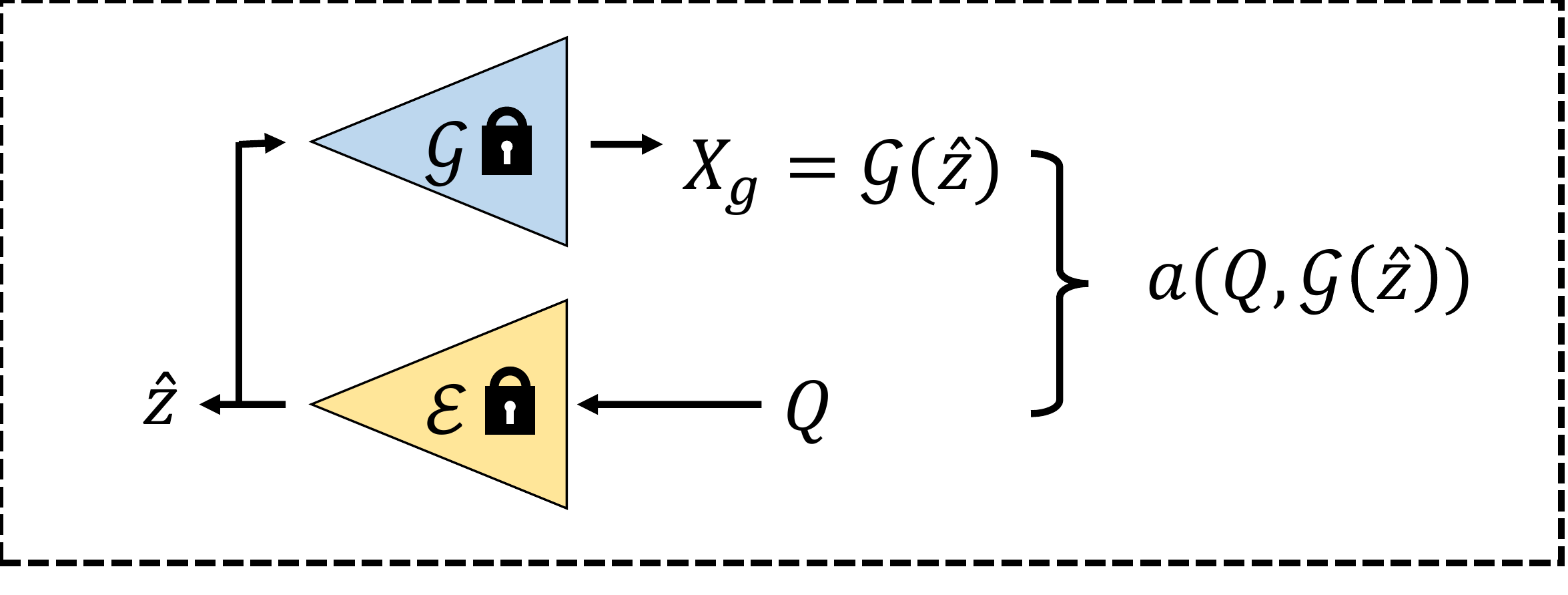}
    \subcaption{Testing}\label{fig:overview_test}
  \end{minipage}
  \caption{An overview of the proposed architecture at (\protect\subref{fig:overview_train}) training and (\protect\subref{fig:overview_test}) testing time. The encoder $\mathcal{E}$ is trained jointly with the generator $\mathcal{G}$. At test time, the discriminator $\mathcal{D}$ is discarded and the parameters of $\mathcal{G}$ and $\mathcal{E}$ are fixed. A query image $Q$ is encoded and compared to its reconstruction $\mathcal{G}(\mathcal{E}(Q))$ in order to find an anomaly score $a$.}
  \label{fig:overview}
\end{figure*}

The architecture of the proposed method is a combination of the progressive growing GAN (pGAN) \cite{Karras2017} and ClusterGAN \cite{Mukherjee2018} but without class labels. An overview of the architecture at both training and testing time is presented in Fig. \ref{fig:overview}. The generator and discriminator are equal to the ones in pGAN \cite{Karras2017}, while the encoder
was inspired by ClusterGAN \cite{Mukherjee2018}. The architecture  and objective function is further described below. At test time, the discriminator is discarded and the parameters of the generator and encoder are fixed. A query image $Q$ is considered to be anomalous or not based on an anomaly score $a$. 

\subsection{Network architecture}\label{sec:architecture}
One of the major drawbacks of AnoGAN \cite{Schlegl2017} is its reliance on accurate reconstruction by a DCGAN \cite{Radford2015}. DCGANs are, among other things, known to suffer from mode collapse \cite{Arjovsky2017}. For that reason, the inventors of AnoGAN replaced the DCGAN with a WGAN-GP \cite{Gulrajani2017} in f-AnoGAN. We instead propose to employ a progressive growing GAN (pGAN) \cite{Karras2017}. pGAN also employs the WGAN-GP loss but incrementally adds new layers to the generator and discriminator while training. This approach has proven to increase the stability and robustness of a GAN, especially in the case of high-resolution images. The generator $\mathcal{G}(z:\theta_G)$ $\mathcal{G}: z\mapsto X_g$ and discriminator $\mathcal{D}(X:\theta_D)$ $\mathcal{D}: X \mapsto Y$ of the proposed method are equal to the ones used in pGAN. The prior, $z \sim \mathcal{N}(0,\mathrm{I}) \in \mathbb{R}^{N_z}$ is drawn from a Gaussian distribution.

Another update in f-AnoGAN compared to AnoGAN was the introduction of an encoder instead of the iterative search, which greatly improved detection speed. The encoder $\mathcal{E}(X:\theta_E)$ maps images to latent space $\mathcal{E}: X \mapsto \hat{z}$. In contrast to f-AnoGAN, the proposed method suggests to train the encoder $\mathcal{E}$ together with $\mathcal{G}$ and $\mathcal{D}$ in the same progressive manner as $\mathcal{G}$ and $\theta_G$ and $\theta_E$ are updated jointly. Various training strategies to learn an encoder have been explored by Dumoulin et. al. \cite{Dumoulin2016}, although on different problems, and they emphasized the importance of learning $\mathcal{G}$ and $\mathcal{E}$ jointly. We make the same observation in our experiments.

Deecke et. al. \cite{Deecke2018} concluded that the discriminator is unsuitable for anomaly detection. While trained to separate real from generated images, thus forcing the two probability distributions to overlap, it is not trained to handle anomalous samples drawn from a different distribution. At test time, see Figure \ref{fig:overview_test}, $\mathcal{D}$ is discarded and the parameters of $\mathcal{G}$ and $\mathcal{E}$, $\theta_G$ and $\theta_E$, are fixed. 

\subsection{Objective function}\label{sec:obj_fun}
Similar to f-AnoGAN and pGAN, we employ the WGAN-GP loss \cite{Gulrajani2017}. However, $\mathcal{E}$ is trained jointly with $\mathcal{G}$, not in a subsequent step as in f-AnoGAN. The GAN objective for the proposed method takes the following form:
\begin{equation}\label{eq:obj_func}
\begin{aligned}
& \min_{\theta_G,\theta_E}\max_{\theta_D} \underset{X\sim p_{\mathrm{data}}}{\mathbb{E}}q(\mathcal{D}(X)) + \underset{z\sim p_z}{\mathbb{E}}q(1-\mathcal{D}(\mathcal{G}(z))) + \\
& \underset{z\sim p_z}{\mathbb{E}}\left\lVert(\mathcal{G}(z)-\mathcal{G}(\mathcal{E}(\mathcal{G}(z))))\right\rVert_1
\end{aligned}
\end{equation}
where $q(x)=x$ since we use a Wasserstein loss \cite{Mukherjee2018}. The third term, $\underset{z\sim p_z}{\mathbb{E}}\left\lVert \mathcal{G}(z)-\mathcal{G}(\mathcal{E}(\mathcal{G}(z)))\right\rVert_1$, is new compared to previous works  \cite{Gulrajani2017,Karras2017,Schlegl2019}.

In contrast to BiGAN and ALI \cite{Dumoulin2016}, the proposed architecture allows $\mathcal{G}$ and $\mathcal{E}$ to interact with each other during training, similar to the encoder used in ClusterGAN. However, while ClusterGAN computes the encoder loss in the latent space $z-\mathcal{E}(\mathcal{G}(z))$, we instead choose to compute the encoder loss in image space $\mathcal{G}(z)-\mathcal{G}(\mathcal{E}(\mathcal{G}(z)))$. The by $\mathcal{G}$ reconstructed query image $Q$ should be the closest match in image space to $Q$ rather than the closest match in latent space, since the anomaly score $a$, see next section, is partly based on a distance measure in image space. Also, the image space loss structures the latent space in a different way than the latent space loss, separating normal and anomalous samples, see the evaluation section.

\subsection{Anomaly detection}\label{sec:anom_det}
We propose to use an anomaly score consisting of two terms, a normalized \textit{residual} and an \textit{origin distance} loss. The \textit{residual} loss $\mathcal{L}_n$ for the query image $Q \in [0,1]^{W\times H \times D}$ is defined as the $\ell_2$-norm between $Q$ and its closest match $\mathcal{G}(\hat{z})$:
\begin{equation}\label{eq:norm_residual}
\mathcal{L}_n(Q,\mathcal{G}(\hat{z})) = \frac{1}{N_X}\left\lVert w(Q)-w(\mathcal{G}(\hat{z}))\right\rVert_2
\end{equation}
where $\hat{z}=\mathcal{E}(Q)$ is the encoded latent vector for image $Q$. In order to minimize the impact of the image contrast to the residual loss, we, unlike f-AnoGAN, propose to apply a minmax normalization $w(x)$ of images. The normalization $w(X) : [\mathrm{min}(X),\mathrm{max}(X)]^{W\times H \times D} \mapsto [0,1]^{W\times H \times D}$ where $\mathrm{min}(X)$ and $\mathrm{max}(X)$ finds the minimum and maximum elements of $X$, is defined as 
\begin{equation}\label{eq:w_x}
w(X)=\frac{X-\mathrm{min}(X)}{\mathrm{max}(X)-\mathrm{min}(X)},
\end{equation}
where the division is element-wise and $N_X=W \cdot H \cdot D$. Without minmax normalization, low contrast samples yield low residual losses and vice versa.

Based on our observations regarding joint encoder and generator training and how that affects the structure of the latent space, we define an \textit{origin distance} loss $\mathcal{L}_o$ as the distance in latent space from encoded vector $\hat{z}$ to the origin:
\begin{equation}
\mathcal{L}_o(\hat{z}) = -\frac{1}{\sqrt{N_z}}\left\lVert\hat{z}\right\rVert_2.
\end{equation}

The anomaly score is then defined as the convex combination between $\mathcal{L}_n$ and $\mathcal{L}_o$ as
\begin{equation}
a(Q,G(\hat{z})) = \lambda\mathcal{L}_n(Q,G(\hat{z}))+(1-\lambda)(\mathcal{L}_o(\hat{z})),
\end{equation}
where $\lambda \in [0,1]$. Samples are classified as anomalies if $a(Q,G(\hat{z})) > \alpha$.

In \cite{Schlegl2019}, f-AnoGAN used a convex combination of a residual loss and a \textit{discrimination} loss as anomaly score. The discrimination loss depends on the difference between the discriminator output and the average discriminator output. In our experiments, adding the discriminator loss did not improve detection results.


\section{Evaluation and results}\label{sec:evaluation}

\subsection{Datasets}
Two different datasets were used for evaluation in this work. The fully annotated KTH-Cellvideos dataset \cite{Gilbert2010,Magnusson2015}, depicting different cells, and the CIFAR-10 dataset \cite{Krizhevsky2012}. All training images were normalized to lie within range $[-1, 1]$.

\subsubsection{CIFAR-10}
The CIFAR-10 dataset \cite{Krizhevsky2012} consists of 50000 $32\times32\times3$ training images in 10 classes (5000 images per class) and 10000 test images (1000 images per class). In this work, a subset of the dataset, denoted as $\mathrm{CIFAR}_{\mathrm{CAR}}$, was used. Images from the car class were treated as normal samples and images from all other classes as anomalous samples. The test set consisted of the 1000 normal test samples (car) and 1000 randomly chosen anomalous test samples from all other classes.

\subsubsection{KTH-Cellvideos}
The KTH-Cellvideos dataset \cite{Gilbert2010,Magnusson2015} consists of grayscale medical images featuring living cells in microscopy image sequences. About 50\% of the labelled objects in the dataset is debris, e.g. bubbles, and they are labelled as such. Events such as mitosis (cell division) and apoptosis (cell death) are also labelled and segmentation masks are available for all cells. In this work, debris is treated as anomalies and cells as normal samples.

The labelled objects in the dataset were split into a training and a test set. All labelled objects (normal/debris) were cropped in a 64 by 64 neighbourhood. In addition, training samples were rotated three times by randomly generated angles. That is, each labelled object (except for the ones reserved for the test set) in the original dataset gave rise to four samples in the training dataset. In total, there were $N = N_n+N_a$ training patches where $N_n=525657$ is the number of normal training patches and $N_a=\frac{\gamma N_n}{1-\gamma}$ the number of anomalous training patches. $\gamma \in[0,1]$ is the user-defined percentage of anomalies in the training data. The test set consisted of 256 normal test images and 256 anomaly test images.



\subsection{Experiments}
In order to evaluate the proposed method, a series of experiments was conducted. Detailed descriptions of network architectures and training configurations are provided in Appendix A. For all experiments, $N_z = 512$ and $\lambda=0.05$. Training of the proposed method was performed on an NVIDIA GTX1080 GPU, the batch size started at 128 and ended at 32 for KTH-Cellvideos and 64 for CIFAR-10. KTH-Cellvideos networks were trained for 48 epochs (6 epochs on full resolution) and CIFAR-10 networks were trained for 32 epochs (4 epochs on full resolution). Training time was about 36 hours for KTH-Cellvideos and about 12 hours for CIFAR-10.

All f-AnoGAN networks were trained with default parameters, batch size 16 and the dimension of z was 128. The KTH-Cellvideos networks were trained for 7 epochs. The training time was about 16 hours for the generator and about 1 hour for the encoder.

The default implementation of f-AnoGAN accepts images of dimension $64\times64\times1$ as input. Images in $\mathrm{CIFAR}_{\mathrm{CAR}}$ have dimension $32\times32\times3$. The default implementation was adapted by increasing the number of channels to 3 and removing one residual block in the discriminator, generator, and encoder respectively.

For dataset $\mathrm{CIFAR}_{\mathrm{CAR}}$, the f-AnoGAN generator was not able to generate visually pleasing images after seven epochs due to the low number of training samples (5000). Even training the network for as much as 70 epochs did not improve the detection performance. Therefore, since more iterations did not improve detection performance, f-AnoGAN was only trained for seven epochs for $\mathrm{CIFAR}_{\mathrm{CAR}}$.
   
Anomaly detection results are measured as the Area Under the Receiver Operating Characteristics (ROC) Curve, abbreviated as AUC \cite{Fawcett2006}.
   


\begin{figure*}[t]
  \centering
  \begin{minipage}[c]{0.66\columnwidth}
    \centering 
    \includegraphics[width=1\columnwidth,trim={4cm 10cm 4cm 10cm},clip]{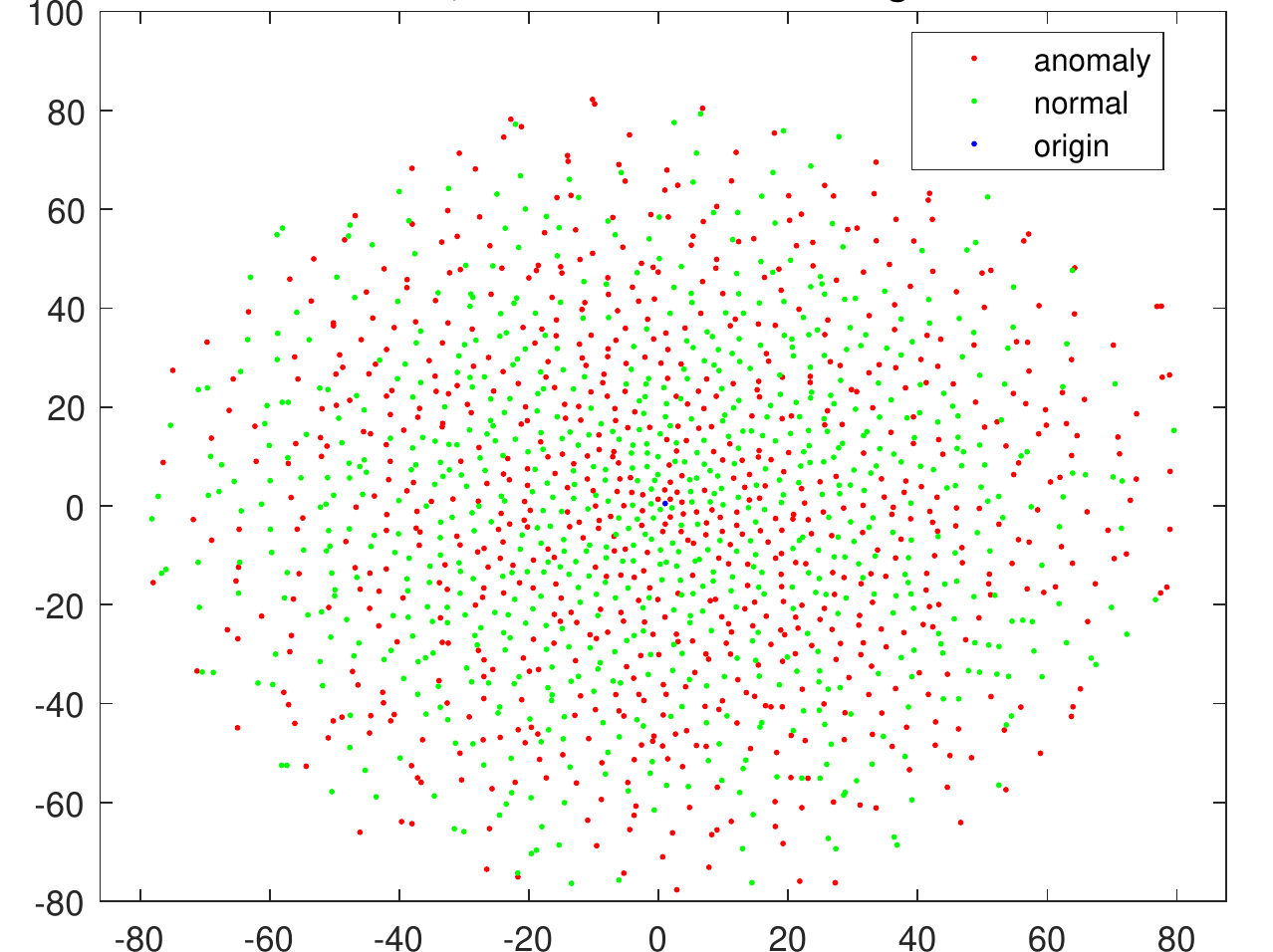}
    \subcaption{0\% Ours, iterative}\label{fig:joint_iterativ}
  \end{minipage}
  \begin{minipage}[c]{0.66\columnwidth}
    \centering 
    \includegraphics[width=1\columnwidth,trim={4cm 10cm 4cm 10cm},clip]{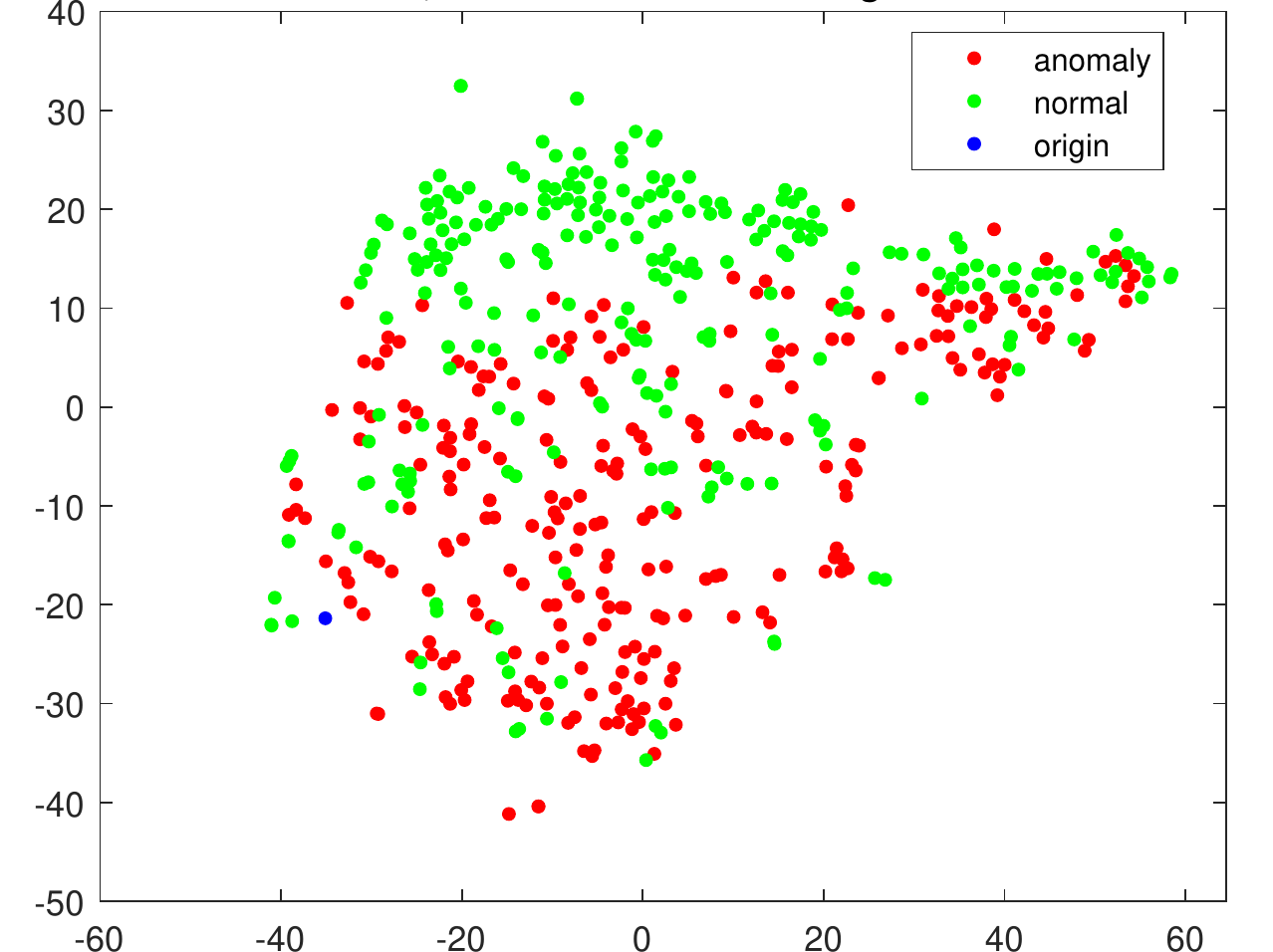}
    \subcaption{0\%, f-AnoGAN}\label{fig:joint_fanogan}
  \end{minipage}
  \begin{minipage}[c]{0.66\columnwidth}
    \centering 
    \includegraphics[width=1\columnwidth,trim={4cm 10cm 4cm 10cm},clip]{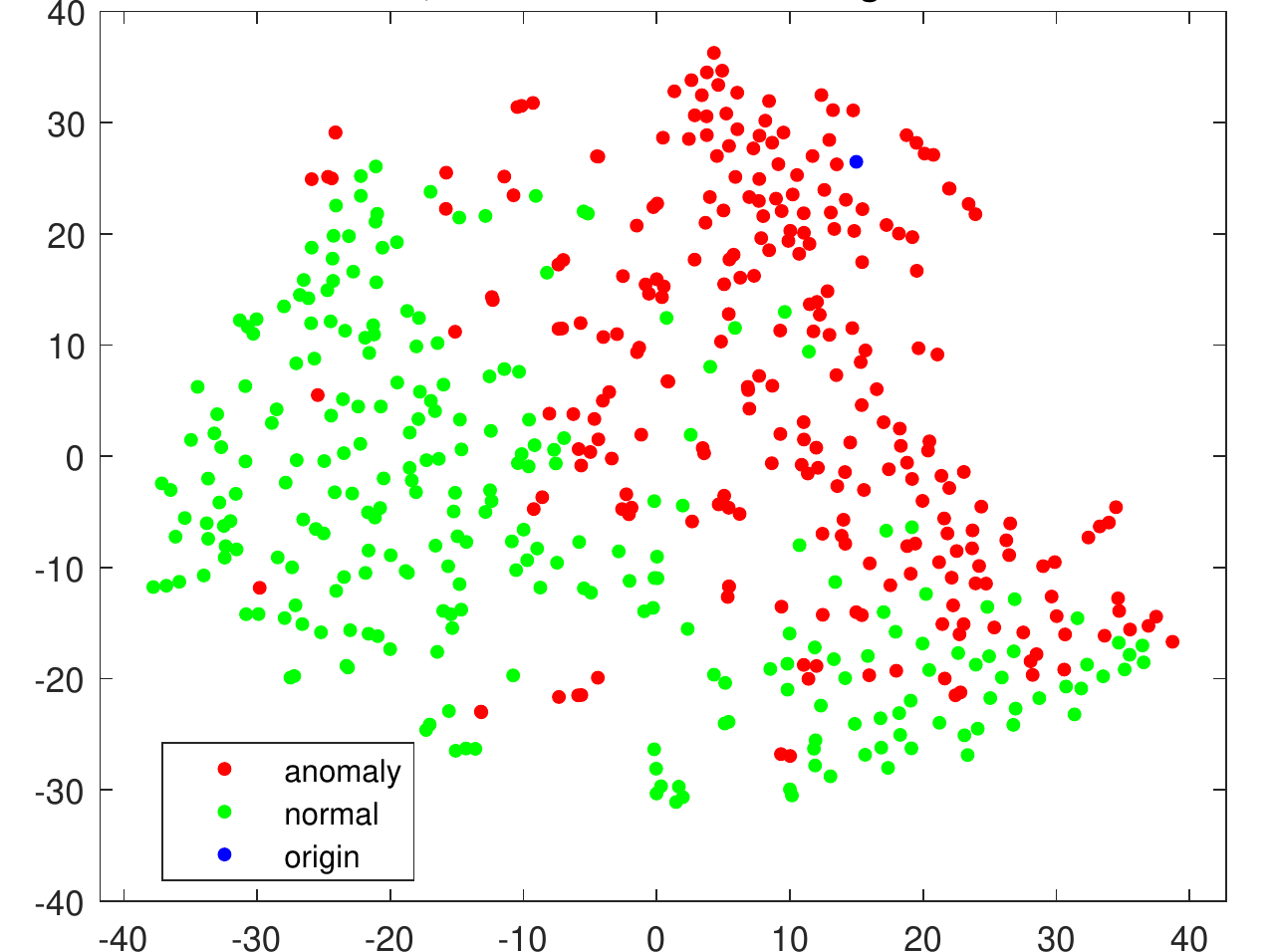}
    \subcaption{0\%, Ours, proposed}\label{fig:joint_proposed}
  \end{minipage}
    \begin{minipage}[c]{0.66\columnwidth}
    \centering 
    \includegraphics[width=1\columnwidth,trim={4cm 10cm 4cm 10cm},clip]{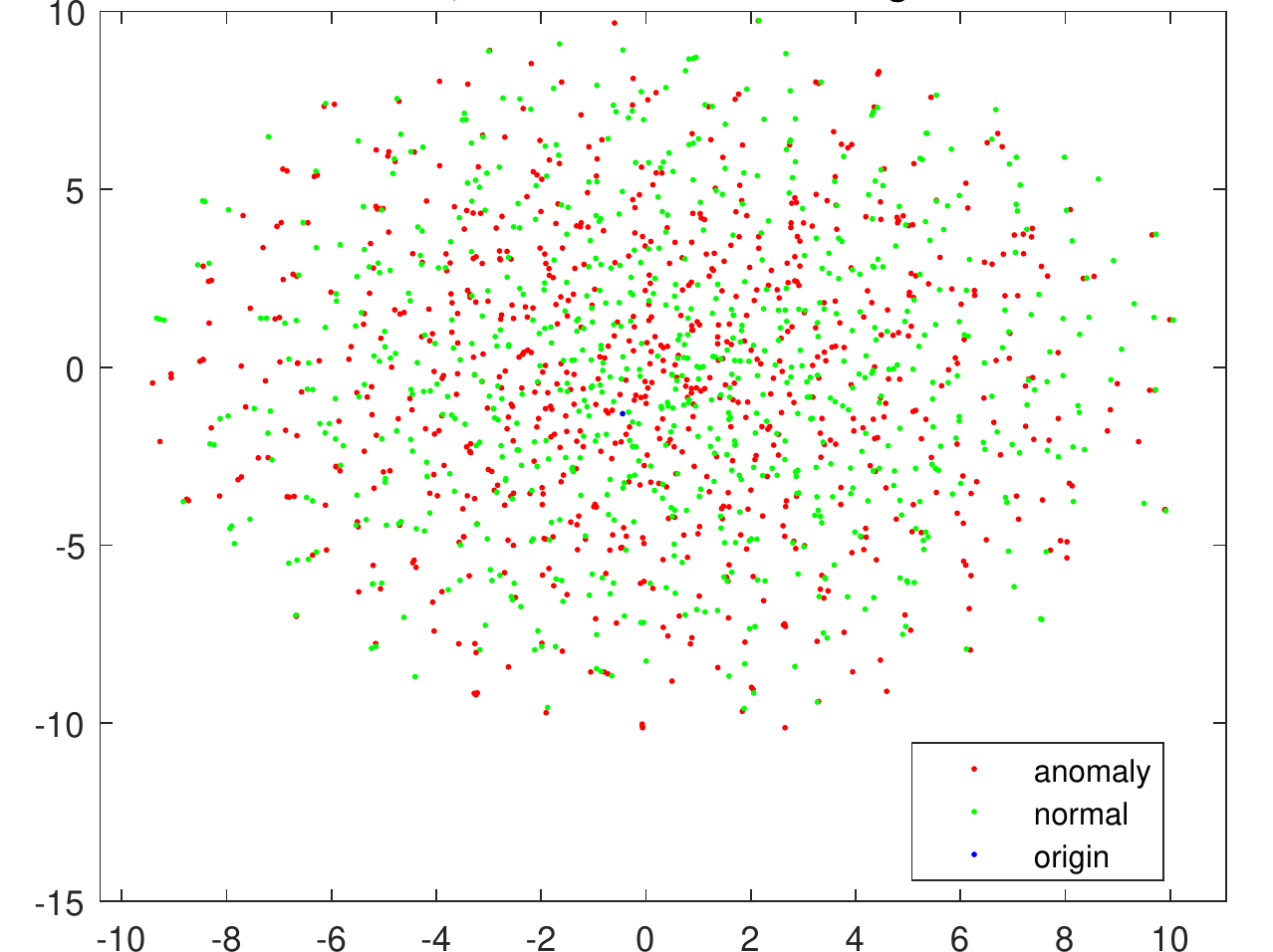}
    \subcaption{2\% Ours, iterative}\label{fig:joint_iterativ_2}
  \end{minipage}
  \begin{minipage}[c]{0.66\columnwidth}
    \centering 
    \includegraphics[width=1\columnwidth,trim={4cm 10cm 4cm 10cm},clip]{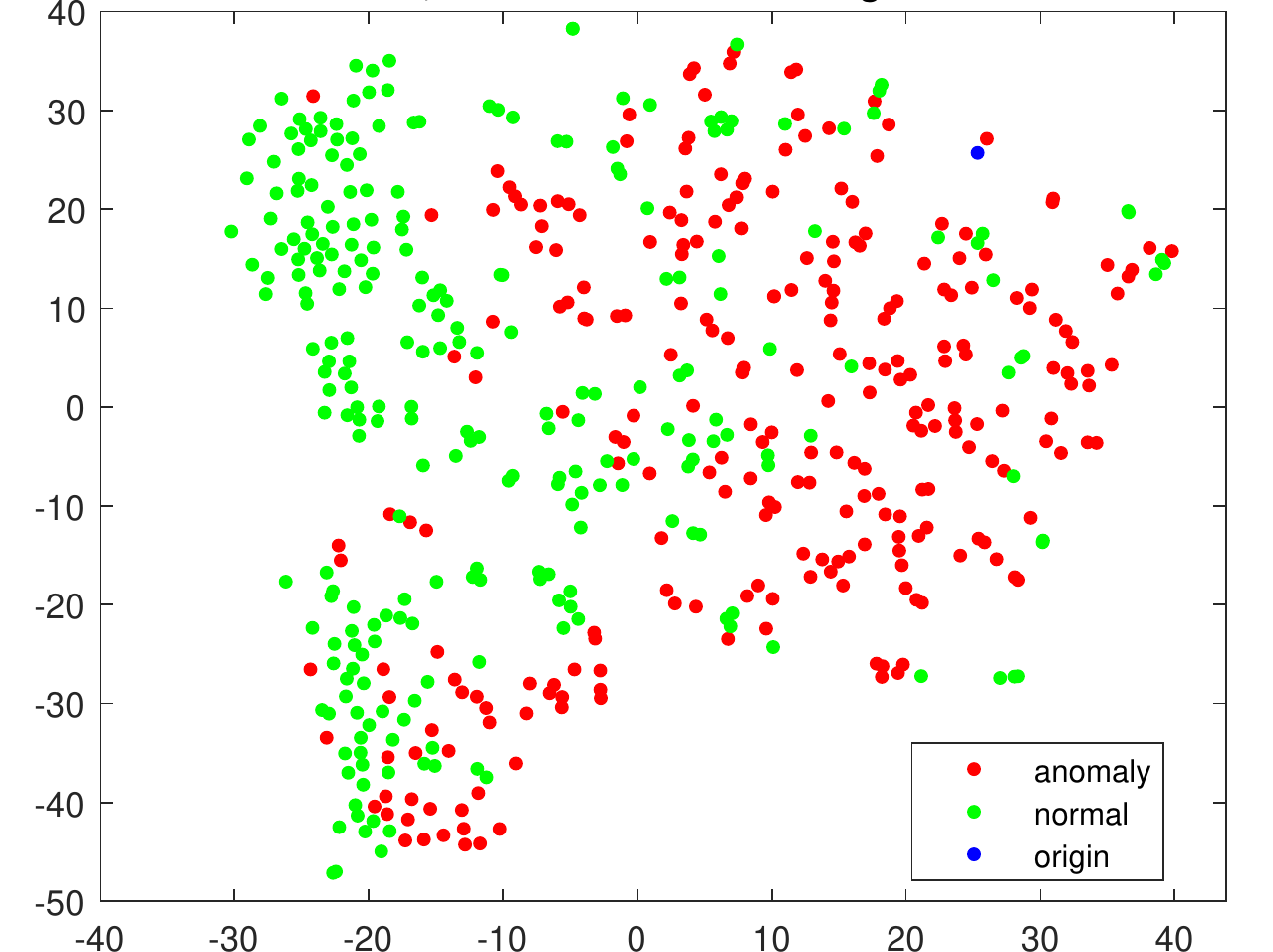}
    \subcaption{2\%, f-AnoGAN}\label{fig:joint_fanogan_2}
  \end{minipage}
    \begin{minipage}[c]{0.66\columnwidth}
    \centering 
    \includegraphics[width=1\columnwidth,trim={4cm 10cm 4cm 10cm},clip]{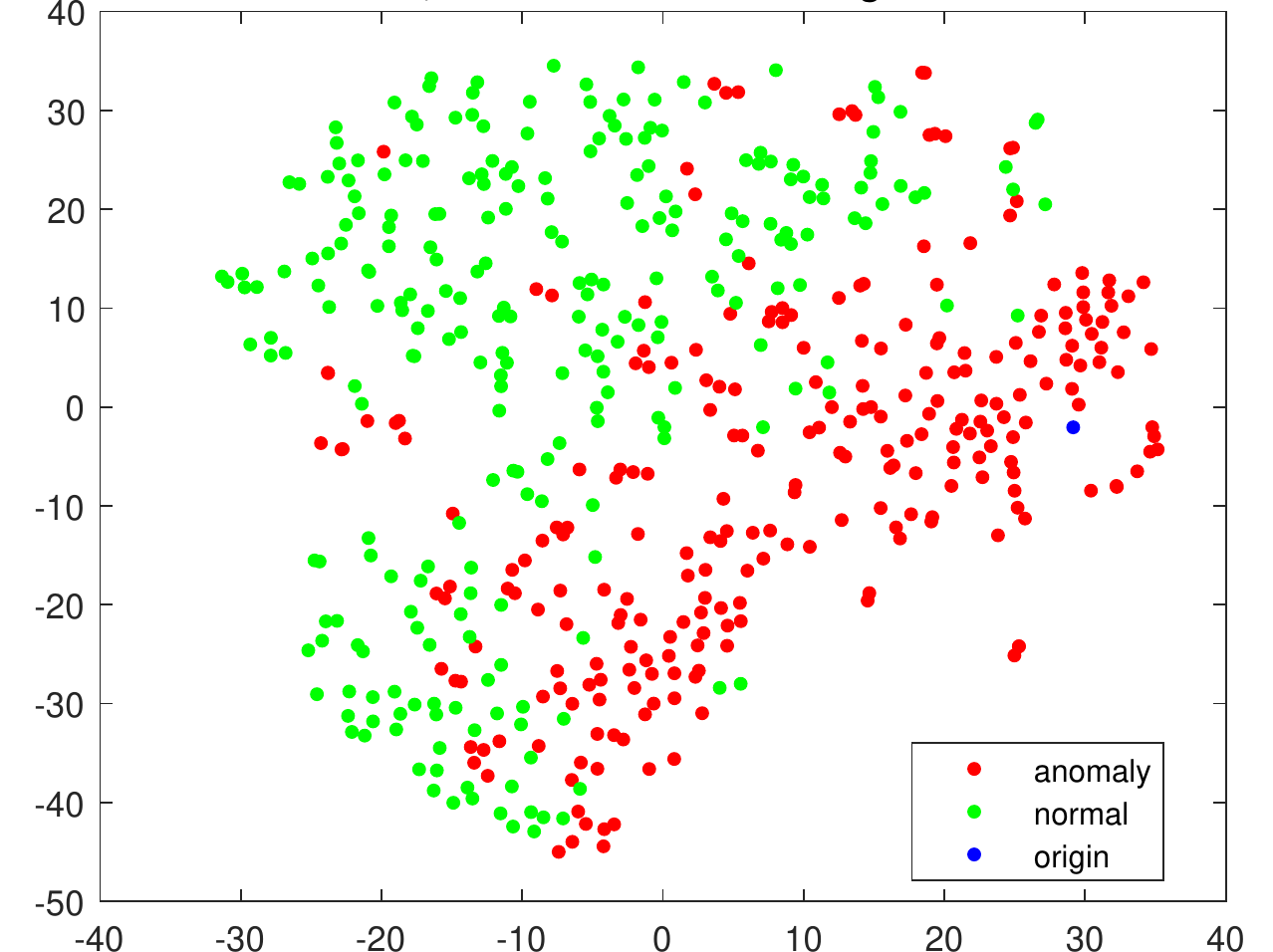}
    \subcaption{2\%, Ours, proposed}\label{fig:joint_proposed_2}
  \end{minipage}
  \caption{t-SNE visualization of validation samples projected to latent space for our method trained (\protect\subref{fig:joint_iterativ},\protect\subref{fig:joint_iterativ_2}) without an encoder and iterative search for closest match, (\subref{fig:joint_proposed},\subref{fig:joint_proposed_2}) with an encoder with latent space projection to find the closest match, and for (\protect\subref{fig:joint_fanogan},\protect\subref{fig:joint_fanogan_2}) f-AnoGAN. The networks were tained on KTH-Cellvideos with (\protect\subref{fig:joint_iterativ}-\protect\subref{fig:joint_proposed}) 0\% and (\protect\subref{fig:joint_iterativ_2}-\protect\subref{fig:joint_proposed_2}) 2\% anomalies in the training data.}
  \label{fig:joint_tsne}
\end{figure*}

\begin{figure*}[]
  \centering
  \begin{minipage}[c]{0.32\textwidth}
    \centering 
    \includegraphics[width=1\textwidth,trim={4cm 10cm 4cm 10cm},clip]{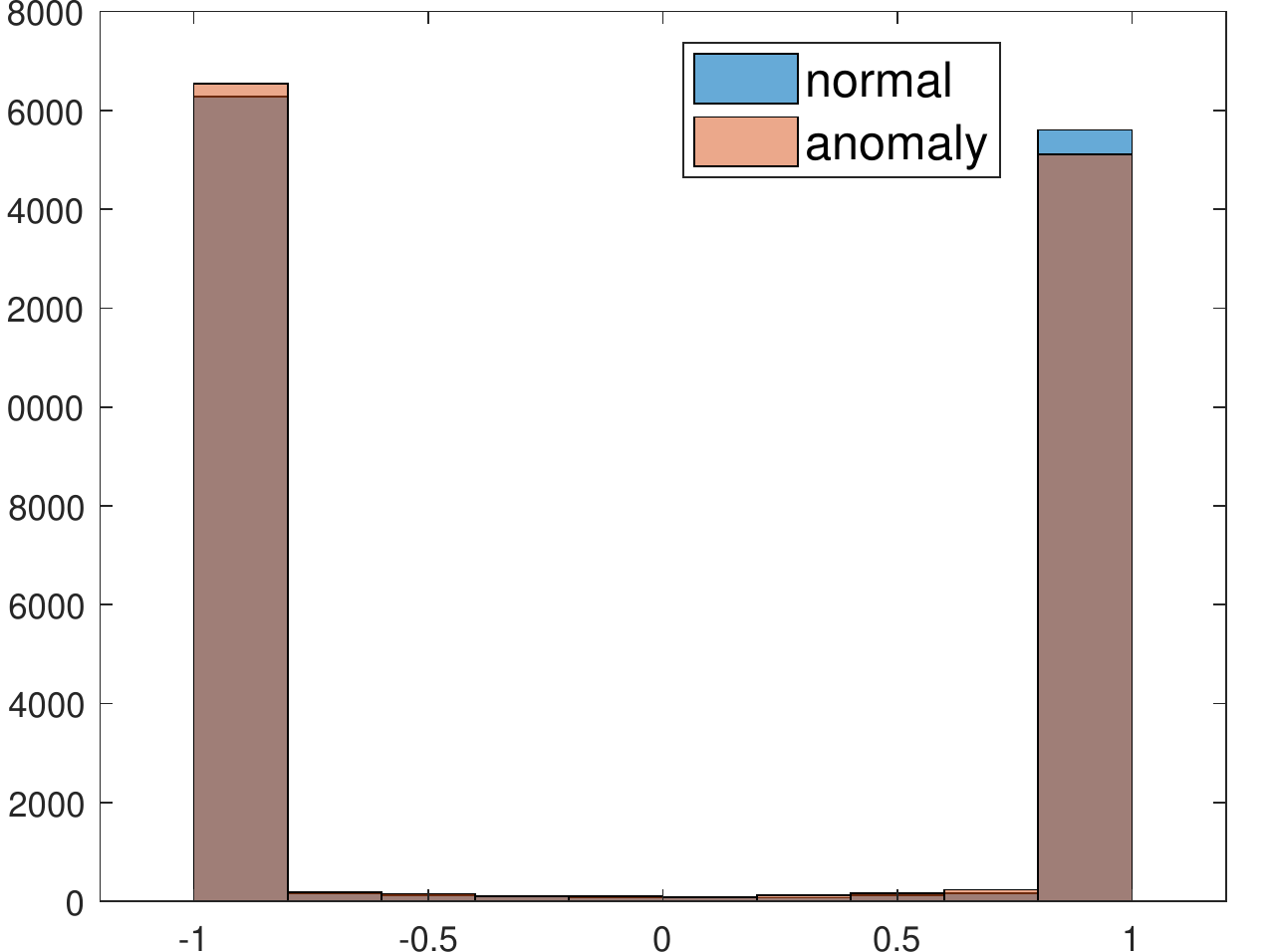}
    \subcaption{f-AnoGAN}\label{fig:coefficients_fanogan}
    \includegraphics[width=1\textwidth,trim={4cm 10cm 4cm 10cm},clip]{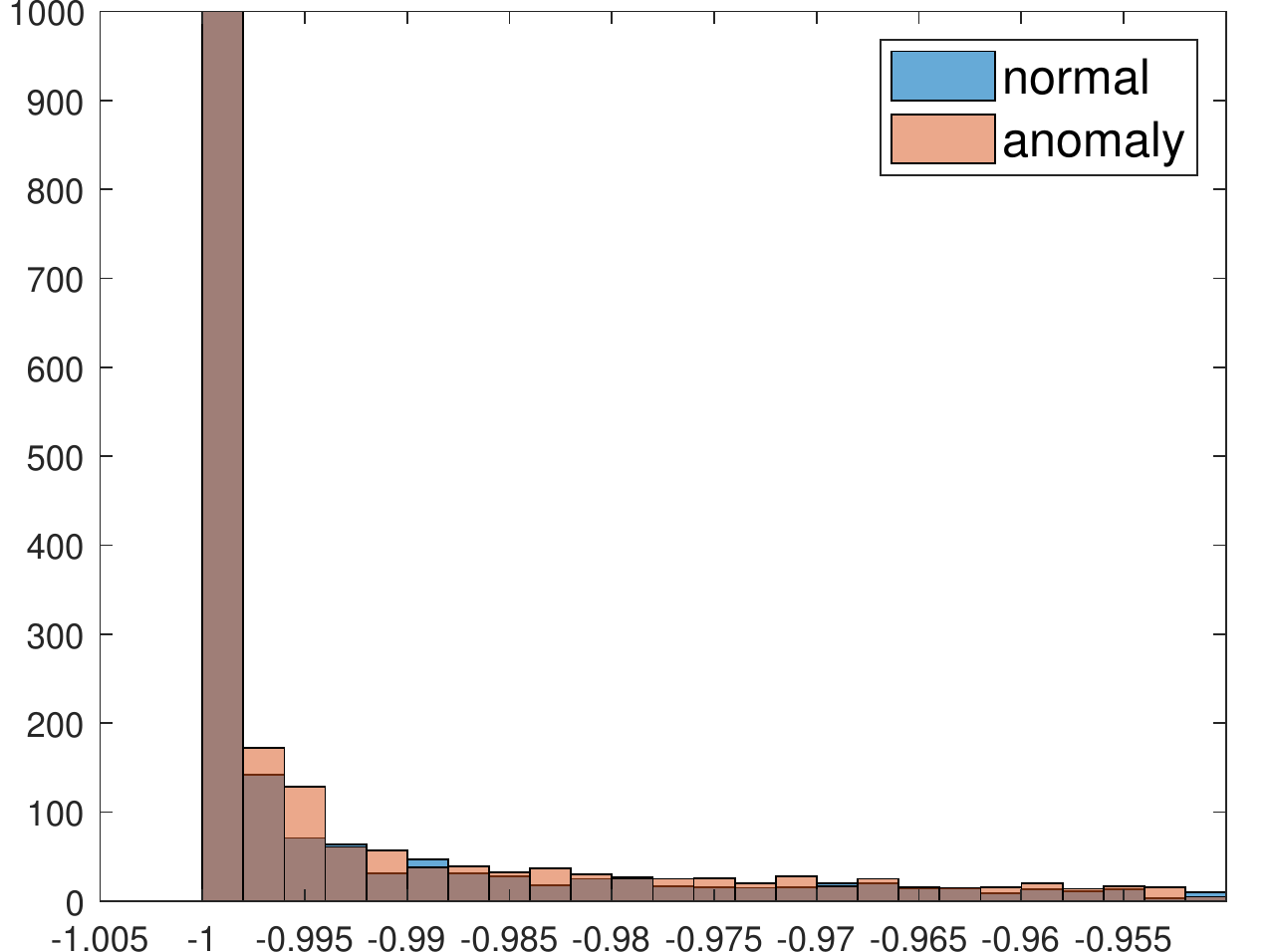}
    \subcaption{f-AnoGAN}\label{fig:coefficients_fanogan_zoom}
  \end{minipage}
  \begin{minipage}[c]{0.65\textwidth}
    \centering 
    \includegraphics[width=1\textwidth,trim={4cm 10cm 4cm 10cm},clip]{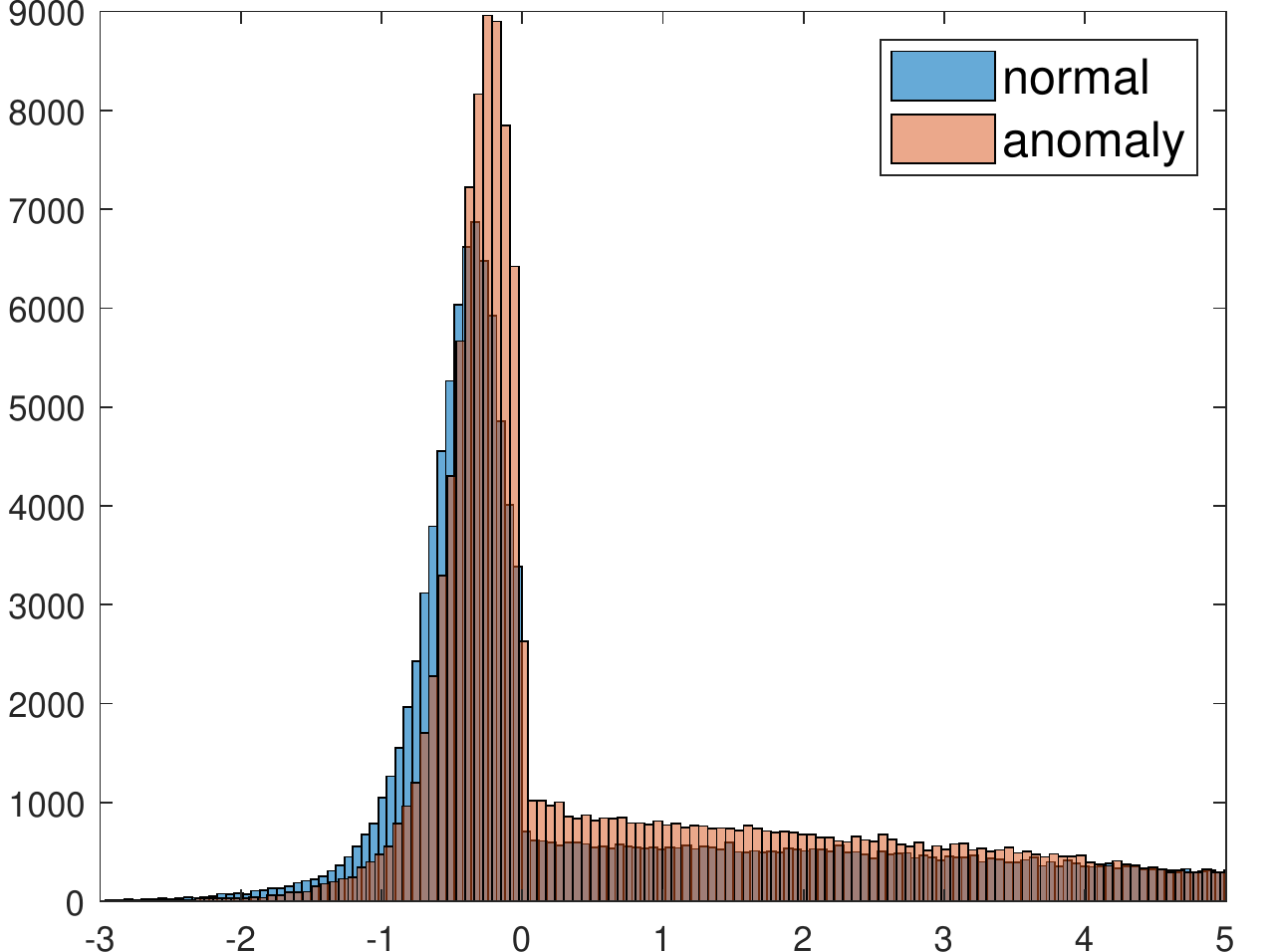}
    \subcaption{Ours}\label{fig:coefficients_ours}
  \end{minipage}
  \caption{Histogram plots for (\protect\subref{fig:coefficients_fanogan}) f-AnoGAN (10 bins) and (\protect\subref{fig:coefficients_ours}) the proposed method (600 bins) of the coefficients of the encoded latent vectors $\hat{z}$ for the validation samples of KTH-Cellvideos. (\protect\subref{fig:coefficients_fanogan_zoom}) shows is a plot of the same data as (\protect\subref{fig:coefficients_fanogan}), but with different axis limits and number of bins (1000 bins).}
  \label{fig:coefficients}
\end{figure*}

\subsubsection{Encoder}\label{sec:encoder_loss}
\paragraph{Training jointly vs. training separately}
In the AnoGAN (note: not f-AnoGAN) paper \cite{Schlegl2017}, an iterative search was used to find the closest match to the query image $Q$ in latent space. The drawbacks with this approach are that a) the optimization can get stuck in local minima, and b) evaluation was time-consuming. Here, we show that when training our method without an encoder and using an iterative search similar to the one in \cite{Schlegl2017}, encoded validation samples lie scattered all over the latent space, see Figure \ref{fig:joint_iterativ}. There is no separation between normal and anomalous samples. 

In contrast, the introduction of an encoder stratifies the latent space. For f-AnoGAN, where the encoder is trained separately, the separation of samples (according to t-SNE) appears to be somewhat worse, Fig. \ref{fig:joint_fanogan}, than for the proposed method, Fig. \ref{fig:joint_proposed}. AUC scores confirming this for the two methods are presented in the anomaly score section below. We believe that the joint encoder training enforces similar images to lie close to each other also in latent space. For the t-SNE plots, a perplexity value of 30 was used and the visualizations were consistent across multiple runs.

In Figure \ref{fig:coefficients}, the histograms of the coefficients of the encoded latent vectors for the validation samples from the KTH-Cellvideo dataset can be found. The networks were trained with 0\% anomalies in the training data. It is clear that the proposed joint encoder training spreads the coefficients more evenly across the latent space, Figure~\ref{fig:coefficients_ours}. These plots also explain why the norm of the latent vector, or the distance to origin, is not a discriminative loss in the case of f-AnoGAN. For f-AnoGAN, the samples end up on a hypercube, Figure~3a-b. In contrast, the density of coefficients is higher for anomalies close to the origin for the proposed method. 

In what follows, we give a possible explanation why the norms of latent variables representing anomalies are empirically smaller than those of normal images. Recall $z \sim \mathcal{N}(0,\mathrm{I}) \in \mathbb{R}^{N_z}$. In the implementation of pGAN, the prior $z$ is normalised to unit length before being processed. A normalized \textit{random} vector $z \in \mathbb{R}^{N_z}$ drawn from $\mathcal{N}(0,\mathrm{I})$ will have small coefficients. GAN training moves data clusters in the latent space away from the origin, otherwise the discriminator would not be able to separate them from the prior distribution, i.e. the noise. The encoder maps normal samples to each cluster respectively. Assuming high intra-class variability among anomalies, anomalies will be mapped away from the clusters and end up closer to the origin, i.e. the noise, and thus have smaller coefficients similar to a \textit{random} vector. 


When the training data is contaminated with anomalies, see Figure \ref{fig:joint_iterativ_2} and \ref{fig:joint_fanogan_2}, the confusion between normal and anomalous samples increases for f-AnoGAN. This is also confirmed in Table \ref{tab:results}, (method d) where the norm-based loss $\mathcal{L}_o$ decreases AUC for f-AnoGAN. In contrast, the proposed method maintains the separability between samples (Figure \ref{fig:joint_proposed_2}) even though the training data is contaminated with as much as 2\% anomalies (method h).

\paragraph*{Distance in image space vs. distance in latent space}
The proposed loss for the encoder is the third term in (\ref{eq:obj_func}), hereby denoted by $d_I$:

\begin{equation}
d_I = \left\lVert \mathcal{G}(z)-\mathcal{G}(\mathcal{E}(\mathcal{G}(z)))\right\rVert_1.
\end{equation}

Generated images $\mathcal{G}(z)$ are compared with their reconstructed images $\mathcal{G}(\mathcal{E}(\mathcal{G}(z))))$ in image space. Another option would be to compare the distance between the latent vector $z$ and the reconstructed latent vector $\hat{z} = \mathcal{E}(\mathcal{G}(z))$ in the latent space:

\begin{equation}
d_z = \left\lVert z-\mathcal{E}(\mathcal{G}(z))\right\rVert_1.
\end{equation}

\begin{figure*}[t]
  \centering
  \begin{minipage}[c]{0.9\columnwidth}
    \centering 
    \includegraphics[width=1\columnwidth,trim={4cm 10cm 4cm 10cm},clip]{figs/dI_tsne.pdf}
    \subcaption{$d_I$}\label{fig:enc_loss_dI}
  \end{minipage}
  \begin{minipage}[c]{0.9\columnwidth}
    \centering 
    \includegraphics[width=1\columnwidth,trim={4cm 10cm 4cm 10cm},clip]{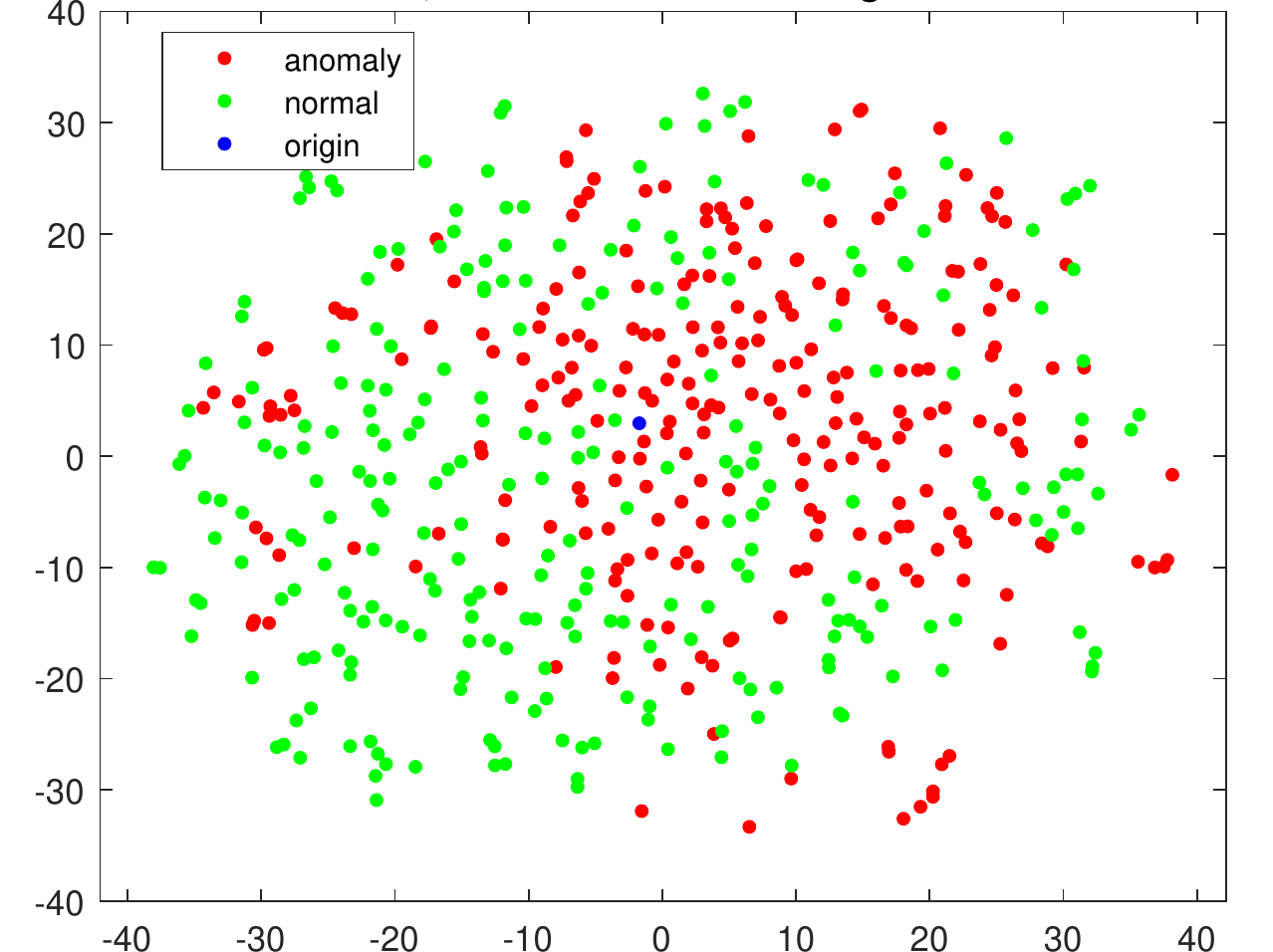}
    \subcaption{$d_z$}\label{fig:enc_loss_dz}
  \end{minipage}
  \caption{t-SNE visualization of validation samples projected to latent space when encoder training loss is based on the distance in (\protect\subref{fig:enc_loss_dI}) image space and (\protect\subref{fig:enc_loss_dz}) latent space.}
  \label{fig:encoder_loss_tsne}
\end{figure*}

Results for the proposed method using $d_I$ and $d_z$ are provided in Table \ref{tab:encoder_loss} and t-SNE visualizations \cite{Maaten2008} of latent space projections are shown in Figure \ref{fig:encoder_loss_tsne}. The network was trained on the KTH-Cellvideos dataset with 0\% anomalies in the training data. Comparing the distance in image space ($d_I$) is clearly preferable when it comes to separation of the validation samples in latent space. A good $d_I$ implies a good $d_z$ but the opposite is not true. We believe this is because $d_I$ enforces similar images (in image space) to lie close to each other also in latent space. Small variations in $z$ and $\hat{z}$ during reconstruction are forced to yield similar images.

\begin{table}
\centering
\caption{AUC results for the proposed method with different encoder losses, $d_z$ and the proposed $d_I$.}
\begin{tabular}{llll}
Encoder loss        & \multicolumn{1}{c}{$\mathcal{L}_n$} & \multicolumn{1}{c}{$\mathcal{L}_o$}  & \multicolumn{1}{c}{$\mathcal{L}_n$ + $\mathcal{L}_o$}\\
\hline
$d_I$ (proposed) & 0.77 & 0.88 & \multicolumn{1}{c}{\textbf{0.90}} \\
$d_z$ & 0.66 & 0.69 & \multicolumn{1}{c}{0.66} \\
\end{tabular}\label{tab:encoder_loss}
\end{table}

\begin{table}
\caption{AUC results for different anomaly losses for the proposed method and f-AnoGAN trained on three different datasets with 0\% and 2\% anomalies.}
\centering
\begin{tabular}{llllllllll}
        &             & \multicolumn{2}{c}{$\mathrm{CIFAR}_{\mathrm{CAR}}$} &  \multicolumn{4}{c}{KTH-Cellvideos}                                           \\
 & Method         & \multicolumn{1}{c}{0\%} & \multicolumn{1}{c}{2\%}&  & \multicolumn{1}{c}{0\%} & \multicolumn{1}{c}{2\%} &  \\
\hline
a) & f-AnoGAN $A$ & 0.45 & 0.44 & & 0.45 & 0.43 \\
b) &f-AnoGAN $\mathcal{L}_r$ & 0.41 & 0.40 & & 0.40 & 0.40 \\
c) &f-AnoGAN $\mathcal{L}_n$ & 0.54 & 0.51 & & 0.78 & 0.76 \\
d) &f-AnoGAN $\mathcal{L}_o$ & 0.53 & 0.50 & & 0.55 & 0.43 \\
e) &Ours $A$ & 0.49 & 0.47 & & 0.55 & 0.53 \\
f) &Ours $\mathcal{L}_r$ & 0.42 & 0.41 & & 0.51 & 0.51 \\
g) &Ours $\mathcal{L}_n$ & 0.58 & 0.56 & & 0.78 & 0.78\\
h) &Ours $\mathcal{L}_o$ & 0.70 & 0.63 & & 0.89 & 0.87 \\
i) &Ours, proposed & \textbf{0.72} & \textbf{0.64} & & \textbf{0.90} & \textbf{0.89}\\
\end{tabular}\label{tab:results}
\end{table}

\newcommand{\resultfigwidth}{0.07}
\newcommand{\queryspace}{0.04mm}
\begin{figure*}[]
  \centering
  \begin{minipage}[c]{0.04\columnwidth}
  	\rotatebox[origin=t]{90}{Q}\vspace{0.2cm}\\
  	\rotatebox{90}{f-AnoGAN}\vspace{0.1cm}\\
  	\rotatebox{90}{Proposed}\vspace{0.05cm}\\
  \end{minipage}
  \begin{minipage}[c]{\resultfigwidth\textwidth}
    \centering 
    \includegraphics[width=1\textwidth]{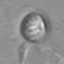}\vspace{\queryspace}
    \includegraphics[width=1\textwidth]{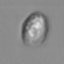}
    \includegraphics[width=1\textwidth]{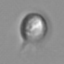}
    \subcaption{}\label{fig:cm:a}
  \end{minipage}
  \begin{minipage}[c]{\resultfigwidth\textwidth}
    \centering 
    \includegraphics[width=1\textwidth]{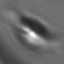}\vspace{\queryspace}
    \includegraphics[width=1\textwidth]{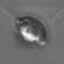}
    \includegraphics[width=1\textwidth]{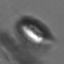}
    \subcaption{}\label{fig:cm:b}
  \end{minipage}
  \begin{minipage}[c]{\resultfigwidth\textwidth}
    \centering 
    \includegraphics[width=1\textwidth]{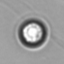}\vspace{\queryspace}
    \includegraphics[width=1\textwidth]{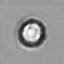}
    \includegraphics[width=1\textwidth]{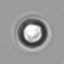}
    \subcaption{}\label{fig:cm:c}
  \end{minipage}
  \begin{minipage}[c]{\resultfigwidth\textwidth}
    \centering 
    \includegraphics[width=1\textwidth]{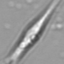}\vspace{\queryspace}
    \includegraphics[width=1\textwidth]{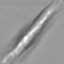}
    \includegraphics[width=1\textwidth]{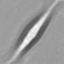}
    \subcaption{}\label{fig:cm:d}
  \end{minipage}
  \begin{minipage}[c]{\resultfigwidth\textwidth}
    \centering 
    \includegraphics[width=1\textwidth]{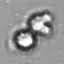}\vspace{\queryspace}
    \includegraphics[width=1\textwidth]{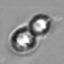}
    \includegraphics[width=1\textwidth]{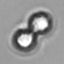}
    \subcaption{}\label{fig:cm:e}
  \end{minipage}
   \begin{minipage}[c]{\resultfigwidth\textwidth}
    \centering 
    \includegraphics[width=1\textwidth]{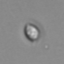}\vspace{\queryspace}
    \includegraphics[width=1\textwidth]{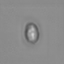}
    \includegraphics[width=1\textwidth]{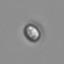}
    \subcaption{}\label{fig:cm:f}
  \end{minipage}
   \begin{minipage}[c]{\resultfigwidth\textwidth}
    \centering 
    \includegraphics[width=1\textwidth]{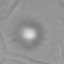}\vspace{\queryspace}
    \includegraphics[width=1\textwidth]{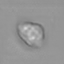}
    \includegraphics[width=1\textwidth]{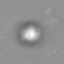}
    \subcaption{}\label{fig:cm:g}
  \end{minipage}
   \begin{minipage}[c]{\resultfigwidth\textwidth}
    \centering 
    \includegraphics[width=1\textwidth]{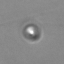}\vspace{\queryspace}
    \includegraphics[width=1\textwidth]{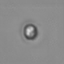}
    \includegraphics[width=1\textwidth]{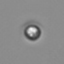}
    \subcaption{}\label{fig:cm:h}
  \end{minipage}
   \begin{minipage}[c]{\resultfigwidth\textwidth}
    \centering 
    \includegraphics[width=1\textwidth]{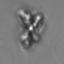}\vspace{\queryspace}
    \includegraphics[width=1\textwidth]{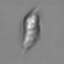}
    \includegraphics[width=1\textwidth]{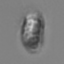}
    \subcaption{}\label{fig:cm:i}
  \end{minipage}
   \begin{minipage}[c]{\resultfigwidth\textwidth}
    \centering 
    \includegraphics[width=1\textwidth]{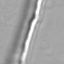}\vspace{\queryspace}
    \includegraphics[width=1\textwidth]{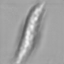}
    \includegraphics[width=1\textwidth]{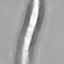}
    \subcaption{}\label{fig:cm:j}
  \end{minipage}
   \begin{minipage}[c]{\resultfigwidth\textwidth}
    \centering 
    \includegraphics[width=1\textwidth]{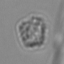}\vspace{\queryspace}
    \includegraphics[width=1\textwidth]{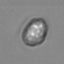}
    \includegraphics[width=1\textwidth]{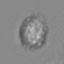}
    \subcaption{}\label{fig:cm:k}
  \end{minipage}
   \begin{minipage}[c]{\resultfigwidth\textwidth}
    \centering 
    \includegraphics[width=1\textwidth]{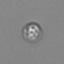}\vspace{\queryspace}
    \includegraphics[width=1\textwidth]{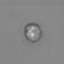}
    \includegraphics[width=1\textwidth]{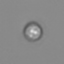}
    \subcaption{}\label{fig:cm:l}
  \end{minipage}
  \caption{Closest matches for query image Q (row 1) by f-AnoGAN (row 2) and the proposed method (row 3). Columns (\protect\subref{fig:cm:a})-(\protect\subref{fig:cm:f}) are examples of cells and columns (\protect\subref{fig:cm:g})-(\protect\subref{fig:cm:l}) are examples of anomalies.}
  \label{fig:closest_matches}
\end{figure*}

\subsubsection{Anomaly score}
As previously described, we propose to use a convex combination of a normalized residual loss $\mathcal{L}_n$ and a norm-based loss $\mathcal{L}_o$. In Table \ref{tab:results}, AUC results for different combinations of these losses for both f-AnoGAN and our method can be seen. The networks were trained on two different datasets with two different percentages of anomalies in the training data. $A$ is the anomaly score proposed in \cite{Schlegl2019} and $\mathcal{L}_r$ is the residual loss, also from \cite{Schlegl2019}, without the proposed minmax normalization. Hence,
\begin{equation}\label{eq:norm_residual}
\mathcal{L}_r(Q,\mathcal{G}(\hat{z})) = \left\lVert Q-\mathcal{G}(\hat{z})\right\rVert_2.
\end{equation}

f-AnoGAN fails to separate normal from anomalous samples in both $\mathrm{CIFAR}_{\mathrm{CAR}}$ and KTH-Cellvideos (see method a and b). Method a) is the default f-AnoGAN implementation. The AUC drastically improves for KTH-Cellvideos when we add the minmax normalization to the residual loss (method c). However, the norm-based loss $\mathcal{L}_o$ cannot discriminate between normal and anomalous samples (method d).

For our method, AUC increases when we add the minmax normalization and the origin distance loss $\mathcal{L}_o$ (method g and h). The proposed method, method i), which uses a convex combination of the two achieves state-of-the art results on both KTH-Cellvideos and $\mathrm{CIFAR}_{\mathrm{CAR}}$.

Regarding training dataset contamination with anomalous samples, there is no degradation in AUC for the proposed method on the dataset KTH-Cellvideos, in contrast to f-AnoGAN. Some examples of closest matches for the proposed method versus f-AnoGAN can be seen in Figure \ref{fig:closest_matches}.

\section{Conclusion}
In this paper, we provide an empirical study of training anomaly detectors using contaminated training data and conclude that detection performance can deteriorate. We also propose an approach to truly unsupervised anomaly detection that can maintain results even when the training data is contaminated with anomalies.

We conclude that \textit{joint} generator and encoder training together with an encoder loss based on image distance is superior to training the encoder and generator separately. Joint generator and encoder training enforces similar images to lie close to each other and, thus, stratifies the latent space. At the same time, robustness to anomalies in the training data is improved.

Further work includes additional analysis of the structure of the latent space and how it is affected by different encoder losses as well as a more extensive study on the choice of the weight $\lambda$. 

\bibliography{nips2019}
\end{document}